\documentclass[11pt,letterpaper]{article}

\usepackage[T1]{fontenc}
\usepackage[utf8]{inputenc}
\usepackage{lmodern}
\usepackage{microtype}
\usepackage[margin=1in]{geometry}
\usepackage{graphicx}
\usepackage{booktabs}
\usepackage{csquotes}
\usepackage[backend=biber,style=apa]{biblatex}
\usepackage[hidelinks]{hyperref}
\hypersetup{
  pdftitle={From Granular Revision Operations to Meaningful Revision Units: Evaluating LLMs for Revision Boundary Detection},
  pdfauthor={Yu Tian, Andrew Potter, Katerina Christhilf,
    Motahareh Darvishpour Ahandani, Jessica Early, Steve Graham,
    Danielle S. McNamara}
}
\title{From Granular Revision Operations to Meaningful Revision Units: Evaluating LLMs for Revision Boundary Detection}
\author{%
Yu Tian\textsuperscript{1,*}, Andrew Potter\textsuperscript{1},
Katerina Christhilf\textsuperscript{1},\\
Motahareh Darvishpour Ahandani\textsuperscript{2},
Jessica Early\textsuperscript{3},\\
Steve Graham\textsuperscript{4},
Danielle S. McNamara\textsuperscript{1}\\[0.8em]
\begin{minipage}{0.92\textwidth}\centering\small
\textsuperscript{1}Learning Engineering Institute, Arizona State University,
Tempe, Arizona, USA\\
\textsuperscript{2}The Polytechnic School, Arizona State University,
Tempe, Arizona, USA\\
\textsuperscript{3}Department of English, Arizona State University,
Tempe, Arizona, USA\\
\textsuperscript{4}Mary Lou Fulton College for Teaching and Learning Innovation,
Arizona State University, Tempe, Arizona, USA\\[0.4em]
\texttt{ytian126@asu.edu; ahpotter@asu.edu; kchristh@asu.edu;
dsmcnama@asu.edu}\\
\texttt{mdarvis2@asu.edu; Jessica.Early@asu.edu; steve.graham@asu.edu}
\end{minipage}
}
\date{}

\begin{document}
\maketitle
\begin{abstract}

Revision traces provide valuable evidence about students' writing processes, but their usefulness for learning analytics depends on how individual revisions are represented. Automated draft-comparison methods often produce granular edit operations that can fragment a single purposeful revision into multiple analytic units. This study evaluates whether LLMs can identify meaningful revision unit boundaries in structured revision operation data and whether they provide value beyond simple non-LLM baselines. Using 113 matched draft--revision pairs from undergraduate writing, expert annotation yielded 4,344 candidate boundaries. We compared zero- and few-shot GPT-5.5 and base Qwen3-32B, parameter-efficient fine-tuning of Qwen3-32B, and majority and proximity-based baselines. Despite receiving revision context and task instructions, no prompted LLM condition outperformed the proximity heuristic (macro-F1 = .825). In contrast, fine-tuned Qwen3-32B using the $\pm 2$ context representation achieved the highest macro-F1 (.859), identifying more same-unit relationships while maintaining precision comparable to the heuristic. Deterministic post-processing substantially improved the prompted models but added little benefit to the strongest fine-tuned model. These findings suggest that LLMs can support revision boundary judgment when task-adapted, but general purpose prompting alone may not outperform transparent structural heuristics.
\end{abstract}


\medskip
\noindent\textbf{Notes for Practice}
\begin{itemize}
  \item Automated revision analysis tools often represent textual changes as granular edit operations, but multiple operations may collectively reflect a single purposeful revision.
  \item General purpose LLM prompting did not outperform a simple proximity heuristic, highlighting the importance of evaluating LLMs against transparent structural baselines.
  \item Task-specific fine-tuning provided additional gains beyond the heuristic, while deterministic post-processing improved non-fine-tuned LLM predictions, suggesting that learned models and transparent task-specific rules can play complementary roles in revision analytics.
\end{itemize}

\section{Introduction}

Revision is a central yet cognitively and procedurally demanding component of writing \parencite[e.g.,][]{conijn2024automated, fitzgerald1987research, flower1986detection}. Writers must continually evaluate the relationship between their intended meaning and the text they have produced, identify problems or opportunities for improvement, and determine how potential changes should be implemented. Whereas a final draft records the endpoint of this process, revisions preserve observable traces of the decisions made during textual development, including changes to ideas, reasoning, organization, wording, and linguistic form. Revision data therefore constitute a valuable source of evidence in learning analytics research for examining writing behavior, feedback uptake, and writing development \parencite{limpo2014children, tian2025exploring, zhu2020effect}.

Computational research on revision based on original and revised drafts commonly extracts revision operations and subsequently classifies their intentions, purposes, or quality \parencite[e.g.,][]{lan2026making, nebel2025towards, zhang2015annotation}. For example, \textcite{jiang2022arxivedits} developed an automated framework for aligning successive versions of scientific manuscripts, extracting fine-grained edits, and classifying the intentions underlying those edits. More recently, TRACE was developed as a framework for representing revisions at multiple levels of granularity through comparisons between source and revised drafts, with the goal of supporting revision annotation and analysis \parencite{tian2026trace}. Such approaches make it possible to analyze revision behavior at scale, but they also raise a fundamental representational question: What should count as one revision?

This question reflects an important distinction between revision operations and revision units. A revision operation describes the formal transformation applied to the text, such as an insertion, deletion, or replacement. A revision unit specifies which operations collectively constitute one purposeful change. Although these concepts are related, an automatically extracted operation does not necessarily represent a complete and purposeful revision. For example, depending on the extraction method, revising \textit{I will see you in a week} to \textit{I will see you in a few weeks} may be represented as the insertion of \textit{few} and the replacement of \textit{week} with \textit{weeks}. Treating these operations as independent revisions would obscure the fact that they jointly serve one semantic purpose: changing the stated time interval. The same problem can arise in more structurally complex revisions. Revising \textit{It is a beautiful day. They decide to go out} to \textit{It is a beautiful day, and they decide to go out} may involve replacing the period with a comma, inserting the conjunction \textit{and}, and changing the initial capitalization of \textit{They}. Although these changes can be represented as separate operations, they collectively realize a single sentence-combining decision.

Fragmentation of this kind has important consequences for learning analytics. Counting each atomic operation as an independent revision may inflate estimates of revision frequency, distort the observed distribution of revision purposes, and result in multiple or conflicting labels being assigned to changes that reflect a single decision. Conversely, imposing a fixed sentence- or paragraph-level unit may combine changes that serve distinct purposes. The challenge is therefore not simply to select a uniformly finer or coarser level of analysis, but to determine contextually which adjacent operations collectively instantiate one purposeful revision.

A small but relevant body of research has addressed an analogous segmentation problem in grammatical error correction \parencite{felice2016automatic, swanson2012correction, xue2014improved}. However, correction detection research has primarily defined a unit as the set of edits required to address one grammatical or lexical error. It has given less attention to broader student revisions in which multiple operations may jointly realize semantic, organizational, stylistic, or rhetorical purposes.

In this study, we use the term \textit{meaningful revision unit (MRU)} to refer to one or more adjacent revision operations that together serve a single revision purpose. We formulate MRU identification as a contextual boundary classification task. Starting from an ordered sequence of TRACE operations, each candidate instance represents the boundary between two adjacent operations. A model determines whether the boundary should be retained because the operations belong to different MRUs or removed because they belong to the same unit. The model is provided with the source and revised sentence contexts as well as neighboring revision operations, allowing the decision to account for the local revision sequence rather than the two focal operations in isolation. This formulation retains the precision of granular operation extraction while enabling related operations to be consolidated into meaningful units.

To determine whether LLM-based reasoning provides value beyond simpler structural information, we evaluate two non-LLM baselines alongside three LLM configurations. The baselines include a majority-class classifier and a proximity-based heuristic that groups adjacent non-preserve operations. The LLM configurations consist of prompting the proprietary GPT-5.5 model, prompting the open-weight Qwen3-32B base model, and adapting Qwen3-32B through parameter-efficient fine-tuning. GPT-5.5 and the base Qwen3-32B model are evaluated under both zero-shot and few-shot prompting, while the fine-tuned Qwen3-32B model is evaluated using an instruction-only prompt. Across the LLM configurations, we examine two local context-window sizes and a deterministic post-processing procedure that enforces a domain-specific structural constraint involving \texttt{preserve} operations.

Through this design, the study makes three contributions. First, it conceptualizes the construction of MRUs as a distinct task situated between granular operation extraction and downstream revision classification. Second, it provides a human-annotated benchmark for evaluating whether adjacent revision operations serve a shared revision purpose. Third, it evaluates whether increasingly sophisticated LLM-based approaches provide value beyond simple structural baselines by systematically comparing majority and proximity heuristics, proprietary and open-weight prompting, parameter-efficient fine-tuning, contextual information, and deterministic post-processing within a common evaluation framework.

\subsection{Research Questions}

The study is guided by the following research questions:

\begin{description}
    \item[\textbf{RQ1.}] How accurately do simple non-LLM baselines, prompted GPT-5.5, prompted base Qwen3-32B, and task-specific fine-tuned Qwen3-32B identify whether two adjacent TRACE operations belong to the same MRU?

    \item[\textbf{RQ2.}] Among the LLM-based approaches, how does MRU boundary-detection performance vary across prompting strategies and local-context configurations?

    \item[\textbf{RQ3.}] For the LLM-based approaches, how does MRU boundary-detection performance change after deterministic post-processing?
\end{description}

Based on prior correction detection research, we expected the proximity heuristic to provide a strong non-LLM baseline because many MRU boundaries may be associated with simple structural patterns in the TRACE operation sequence. At the same time, because some MRU decisions require interpreting how adjacent operations function in context, we expected the strongest task-specific fine-tuned model to outperform the simple baselines by identifying same-unit relationships that cannot be captured from operation type alone. Among the LLM-based approaches, we expected the fine-tuned Qwen3-32B model to outperform prompted models, given prior evidence that task-specific adaptation can improve performance on specialized text classification and revision analysis tasks. We also expected few-shot prompting to improve performance relative to zero-shot prompting by providing concrete examples of the target boundary judgments. Because MRU decisions may depend on neighboring operations, we expected the larger $\pm 2$ context window to support stronger performance than the $\pm 1$ window. Finally, because the post-processing rule encoded an annotation constraint involving \texttt{preserve} operations, we expected post-processing to increase precision for LLM-based predictions by reducing invalid same-unit predictions. Comparisons between GPT-5.5 and the base Qwen3-32B model were treated as exploratory because prior work did not provide a clear directional basis for predicting which prompted model would perform better on this task.

\section{Related Work}

A particularly relevant line of research concerns grammatical correction detection and merging, where the goal is to determine which low-level edit operations collectively constitute a single correction. This problem arises because automatic comparison of an original sentence with its corrected version typically produces atomic operations such as insertions, deletions, and substitutions, whereas a single human correction may require multiple such operations. Consequently, grouping related edits into complete correction units is an important prerequisite for subsequent error classification and analysis.

\textcite{swanson2012correction} provided an early computational treatment of this problem in a system for analyzing corrections made by teachers to ESL essays. Their framework first compared source and corrected sentences using minimum edit distance, producing sequences of insertion, deletion, substitution, and unchanged operations. To determine which nonmatching operations belonged to the same correction, they evaluated a family of proximity-based heuristics that differed in how many unchanged words could occur between edits before the edits were treated as separate corrections. Their strongest heuristic merged consecutive nonmatching operations when no unchanged word intervened. This strategy produced more accurate correction boundaries than either treating every atomic edit as an independent correction or merging edits across larger distances.

\textcite{xue2014improved} extended this line of work by reformulating edit merging as a supervised binary classification task. After extracting basic insertions, deletions, and substitutions, they created one classification instance for each pair of consecutive edits. A pair was labeled \texttt{True} when the two edits belonged to the same expert-annotated correction and \texttt{False} otherwise. A maximum-entropy classifier then predicted whether the two edits should be merged using features derived from the edits and their surrounding linguistic contexts. This approach treated correction-boundary detection as a context-sensitive classification problem rather than relying only on edit proximity. Across multiple ESL corpora and annotation schemes, the learned merging model improved correction detection relative to heuristic merging approaches.

\textcite{felice2016automatic} addressed the same problem by incorporating richer linguistic information into both alignment and merging. They developed a linguistically enhanced alignment algorithm to identify more plausible mappings between source and corrected tokens and to derive a sequence of token-level edit operations. They then applied a recursive, rule-based merging procedure to determine which operations should be combined into complete learner-error spans. The merging rules captured linguistically motivated patterns involving punctuation and capitalization, possessives, whitespace changes, phrasal expressions, part-of-speech relationships, and multi-token lexical changes. Their method outperformed simpler alternatives that either treated all operations separately, merged all consecutive nonmatches, or merged edits solely on the basis of operation type. The linguistically informed extraction approach developed by Felice et al.\ was later incorporated into ERRANT, which automatically extracts edits and assigns grammatical-error categories at multiple levels of granularity \parencite{bryant2017automatic}.

Together, these grammatical error correction studies demonstrate that atomic edit operations do not necessarily constitute appropriate units for downstream analysis. They also illustrate several strategies for constructing higher-level units from low-level edits, including proximity-based heuristics, learned pairwise classification, and linguistically informed rule-based merging. However, these approaches were developed primarily for grammatical error correction, where edits are grouped because they jointly repair the same grammatical, lexical, or mechanical error. The revision unit identification problem addressed in the present study is broader. In student writing, multiple adjacent operations may belong to the same unit not because they correct the same error, but because they jointly realize a shared communicative or rhetorical purpose. Determining whether such operations should be grouped therefore requires reasoning about their function in context rather than relying only on proximity or predefined linguistic patterns.

A related representational challenge also appears in writing process analytics based on keystroke data. Keystroke logs capture highly granular actions, but these actions often need to be aggregated into larger revision events before they can be meaningfully interpreted. For example, \textcite{conijn2024automated} developed methods for reconstructing revision events from granular keystroke data, while \textcite{mouchel2023understanding} examined revision behavior as an indicator of learner engagement and self-regulated writing activity. Although this line of research operates directly on process data, it reflects the same broader measurement problem: low-level computational traces must be transformed into units that more closely correspond to meaningful learner actions before they can support interpretable learning analytics. As in revision unit construction from text comparisons, this aggregation requires contextual judgments about how neighboring actions in the writing process relate to one another and whether they jointly constitute a coherent revision event.

Recent advances in large language models (LLMs) provide a promising way to support these context-sensitive judgments. LLMs have already been applied to related revision classification tasks. \textcite{ruan2024large}, for example, investigated prompting and fine-tuning approaches for classifying the intentions underlying edits in scientific document revisions, while \textcite{liu2025efficient} examined parameter-efficient fine-tuning for revision-intention prediction. These studies assume that the revisions to be classified have already been delineated and therefore do not directly address revision unit boundaries. Nevertheless, they provide evidence that LLMs can be adapted to make nuanced judgments about the relationship between original and revised text. The present study extends this capability to an earlier stage of the revision analysis pipeline. Rather than assigning a purpose to a predefined revision, we investigate whether LLMs can determine which adjacent granular revision operations should be grouped into the same MRU and, importantly, whether this contextual modeling provides value beyond simpler structural heuristics.

\section{Methods}

\subsection{Data Source}

The data were drawn from a larger study examining the effects of different prompting strategies for AI-generated writing feedback on student writing outcomes \parencite{tian-inprep-feedback}. Participants were 135 undergraduate students enrolled at a public university in the United States. During the first session of an orientation program, students were given 30 minutes to write a narrative essay about a pivotal moment in their lives. Approximately one month later, during a subsequent session of the same program, students were asked to revise their original essays in response to AI-generated feedback within a 20-minute period. Of the 135 students who participated in the initial writing session, 113 completed the revision task, yielding 113 matched draft--revision pairs and an attrition rate of 16.30\%.

\subsection{TRACE Revision Analysis}

The 113 draft--revision pairs were analyzed using TRACE, a tool for estimating interpretable revision operations from source--target draft pairs \parencite{tian2026trace}. TRACE identifies five operation types (preserve, insert, delete, replace, and move) and represents the resulting operations in tabular form according to their sequence in the source and revised texts. This representation provides a granular account of how the source draft was transformed into the revised version and supports subsequent annotation of higher-level revision units. Across the 113 draft--revision pairs, TRACE identified 4,457 operation rows. Table~\ref{tab:trace-operations} presents the distribution of these operations.

\begin{table}[htbp]
\centering
\caption{Instance Counts for TRACE Revision Operations}
\label{tab:trace-operations}
\begin{tabular}{lr}
\toprule
\textbf{Operation} & \textbf{Count} \\
\midrule
Preserve & 2,390 \\
Replace & 1,015 \\
Insert & 646 \\
Delete & 405 \\
Move & 1 \\
\midrule
\textbf{Total} & \textbf{4,457} \\
\bottomrule
\end{tabular}
\end{table}

\subsection{Meaningful Revision Unit Annotation}

Four trained expert annotators examined the TRACE operation sequences to identify MRUs. All four annotators had backgrounds in education; two held doctoral degrees and two were doctoral candidates.

\begin{figure}[htbp]
    \centering
    \includegraphics[width=0.70\textwidth]{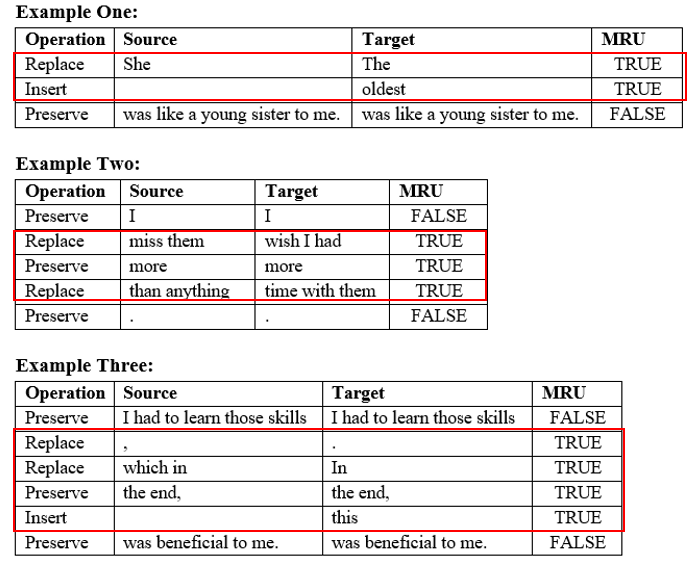}
    \caption{Three examples of MRUs identified in the dataset, illustrating typical variation in their scale and internal structure.}
    \label{fig:revision-unit-examples}
\end{figure}

In this study, an MRU was defined as a purposeful modification to the text that served a single revision purpose. Such purposes could range from relatively local changes, such as typographical correction or vocabulary substitution, to more complex changes involving sentence restructuring or discourse-level reorganization. Accordingly, MRU identification was not determined solely from operation type or textual proximity. Instead, annotators considered how adjacent operations functioned together in the surrounding source and revised text. Figure~\ref{fig:revision-unit-examples} illustrates how MRUs may vary in linguistic scale and internal structure.

In Example One, replacing \textit{She} with \textit{The} and inserting \textit{oldest} jointly form a single lexical modification, while the following preserved text falls outside that revision unit. Example Two illustrates a more complex semantic revision: replacing \textit{miss them} with \textit{wish I had}, preserving \textit{more}, and replacing \textit{than anything} with \textit{time with them} collectively transform the meaning of the sentence and are therefore treated as one MRU. This example also illustrates an important annotation rule concerning \texttt{preserve} operations. A \texttt{preserve} operation may occur within an MRU when it connects surrounding revision operations that jointly serve the same purpose, but it cannot independently begin or end an MRU. Thus, the preserved word \textit{more} is included because it occurs between two related revision operations that together realize the same semantic change. Example Three shows a sentence-restructuring revision in which punctuation replacement, modification of \textit{which in} to \textit{In}, preservation of \textit{the end,}, and insertion of \textit{this} collectively realize a coordinated restructuring of the sentence. Together, these examples show that an MRU may consist of a single operation or a sequence of heterogeneous operations whose relationship becomes apparent only when considered in context.

Annotation was conducted at the boundary level. For each pair of consecutive TRACE operations within a draft--revision pair, annotators made a binary decision indicating whether the two operations should be combined into the same MRU. A \texttt{True} decision indicated that the adjacent operations belonged to the same unit, whereas a \texttt{False} decision indicated a boundary between two units. Because a sequence containing $n$ TRACE operations contains $n - 1$ candidate boundaries, the 4,457 operation rows across 113 draft--revision pairs yielded 4,344 candidate boundary decisions. The TRACE files were approximately evenly distributed among the four annotators, with each file independently annotated by two annotators.

Inter-rater reliability was calculated on the 4,344 candidate boundaries before adjudication. Because rater pairings varied across TRACE files, agreement statistics were calculated by pooling all doubly coded boundary decisions across annotator pairs. Overall, the paired annotators agreed on 94.28\% of the boundary decisions. Given the substantial class imbalance and the sensitivity of Cohen's kappa to prevalence, Gwet's AC1 was used as the primary chance-corrected agreement statistic \parencite{vach2023gwet}. The resulting agreement coefficient was AC1 = .91, indicating very high agreement beyond chance. Cohen's unweighted kappa was also calculated as a complementary measure and yielded $\kappa = .85$, likewise indicating a high level of chance-corrected agreement.

Disagreements between the two annotators were subsequently reviewed and resolved by a more experienced expert adjudicator. The adjudicated annotations constituted the gold-standard labels used for model development and evaluation.

\subsection{Boundary Instance Construction and Data Partitioning}

The adjudicated TRACE revision data were transformed into boundary classification instances. Each instance corresponded to one candidate boundary between two consecutive TRACE operations. The prediction target was whether the two focal operations belonged to the same MRU. A positive label (\texttt{same\_unit = True}) indicated that the operations should be grouped into the same unit, whereas a negative label (\texttt{same\_unit = False}) indicated that they should remain separated by a revision unit boundary.

\begin{figure}[htbp]
    \centering
    \includegraphics[width=0.78\textwidth]{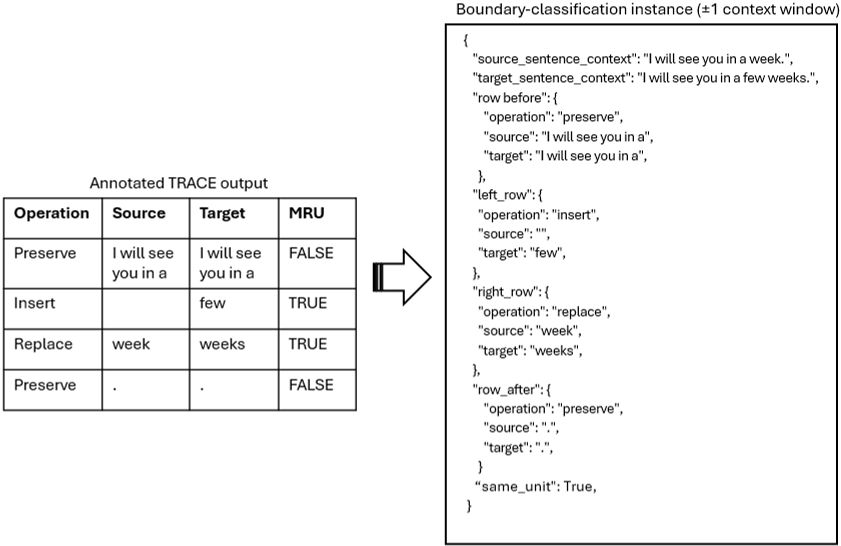}
    \caption{Example transformation of annotated TRACE revision operations into a boundary classification instance using a $\pm 1$ context window. The focal pair consists of the two adjacent operations to be classified, with one preceding and one following operation included as local context.}
    \label{fig:data-transformation}
\end{figure}

Each boundary instance contained the two focal TRACE operations, represented as a left operation and a right operation, together with the corresponding source- and target-sentence contexts. To examine whether additional local revision context supported boundary decisions, two contextual window sizes were constructed. In the $\pm 1$ condition, the instance additionally included the TRACE operation immediately preceding and immediately following the focal pair, when available. In the $\pm 2$ condition, up to two preceding and two following operations were included. Thus, the two conditions differed in the amount of neighboring revision-operation context available to the model. Figure~\ref{fig:data-transformation} illustrates the transformation from annotated TRACE output to a boundary classification instance using the $\pm 1$ context window.

The data were divided into training, validation, and test sets using an approximately 80/10/10 allocation. To prevent information from the same writing sample from appearing in multiple subsets, partitioning was conducted at the draft-pair level rather than at the individual boundary instance level. A fixed random seed of 42 was used to ensure reproducibility of the partitioning. The resulting subsets contained 3,532 training instances, 371 validation instances, and 441 test instances, corresponding to approximately 81.3\%, 8.5\%, and 10.2\% of the full dataset, respectively. Table~\ref{tab:data-partitions} summarizes the class distribution across the three subsets.

\begin{table}[htbp]
\centering
\caption{Distribution of Draft Pairs and Boundary Instances Across Data Partitions}
\label{tab:data-partitions}
\begin{tabular}{lrrrr}
\toprule
\textbf{Partition} & \textbf{Draft pairs} & \textbf{False} & \textbf{True} & \textbf{Total} \\
\midrule
Training & 90 & 2,930 & 602 & 3,532 \\
Validation & 11 & 315 & 56 & 371 \\
Test & 12 & 364 & 77 & 441 \\
\midrule
\textbf{Total} & \textbf{113} & \textbf{3,609} & \textbf{735} & \textbf{4,344} \\
\bottomrule
\end{tabular}
\end{table}

The test set was held out from model development and used only for final evaluation. Examples used for few-shot prompting of GPT-5.5 and the base Qwen3-32B model were drawn exclusively from the training set.

\subsection{Non-LLM Baselines}

Two simple non-LLM baselines were included to contextualize LLM performance and determine how much MRU boundary structure could be recovered without linguistic or semantic modeling. First, a majority-class baseline predicted \texttt{same\_unit = False} for every candidate boundary, corresponding to the majority class in the data. Because the boundary labels were substantially imbalanced toward \texttt{False}, this baseline provided a reference for interpreting accuracy and macro-F1.

Second, a proximity-based heuristic was implemented following the general logic of correction detection approaches that merge adjacent edits when no unchanged material intervenes \parencite{swanson2012correction}. The heuristic predicted \texttt{same\_unit = True} when both focal TRACE operations were non-preserve revision operations (i.e., \texttt{insert}, \texttt{delete}, \texttt{replace}, or \texttt{move}). If either focal operation was \texttt{preserve}, the heuristic predicted \texttt{same\_unit = False}. The heuristic used only operation type and immediate adjacency and did not use source or revised text, surrounding operation context, labeled training data, or model-based reasoning. It therefore provided a direct test of how strongly simple structural regularities in the TRACE representation predicted MRU boundaries.

\subsection{LLM-Based Boundary Judgement}

The modeling experiments examined whether LLMs could identify MRUs from the boundary instances constructed from TRACE operations. Three model configurations were evaluated: (a) GPT-5.5, a proprietary LLM evaluated through prompting; (b) the base Qwen3-32B open-weight model without task-specific fine-tuning; and (c) Qwen3-32B adapted to the boundary classification task through parameter-efficient fine-tuning.

GPT-5.5 and the base Qwen3-32B model were each evaluated under zero-shot and few-shot prompting conditions. The fine-tuned Qwen3-32B model was evaluated using an instruction-only prompt because it had already been adapted to the task using labeled training instances. All model configurations were evaluated separately using the $\pm 1$ and $\pm 2$ boundary instance formats described above. At inference, model responses included a binary \texttt{same\_unit} prediction together with structured information about the candidate boundary, confidence, and a brief evidence-based linguistic explanation. Only the \texttt{same\_unit} prediction was used in the performance analyses. LLM predictions were evaluated both before and after the deterministic post-processing procedure described in Section~\ref{sec:postprocessing}.

\subsubsection{GPT-5.5 Prompting}

GPT-5.5 was evaluated using zero-shot and few-shot prompting. In both conditions, the model was instructed to assume the role of an expert annotator of writing revisions and was provided with the operational definition of an MRU. For each candidate boundary, the model was asked to return structured JSON containing \texttt{candidate\_boundary\_id}, a binary \texttt{same\_unit} prediction, a confidence value between 0 and 1, and a brief evidence-based \texttt{linguistic\_analysis}.

The zero-shot prompt contained the task description, definition of an MRU, decision criteria, contextual information, and required output format but did not include labeled demonstrations. The few-shot prompt used the same instructions and output requirements but additionally included four labeled examples drawn from the training set. The same four demonstrations were used for every test instance. Two represented positive (\texttt{True}) cases and two represented negative (\texttt{False}) cases. Within each class, one example represented a relatively straightforward boundary decision and the other represented a more difficult decision. This selection was intended to expose the model to both outcome classes and to variation in task difficulty while holding the demonstrations constant across test instances.

Both prompting conditions were evaluated separately using the $\pm 1$ and $\pm 2$ boundary instance formats. GPT-5.5 was queried using the model identifier \texttt{gpt-5.5}, with temperature set to 0 to reduce sampling variability. No explicit maximum output-token limit was imposed. All 441 test instances produced valid outputs from which the binary predictions could be extracted.

\subsubsection{Base Qwen3-32B}

The base Qwen3-32B model (\texttt{Qwen/Qwen3-32B}) was evaluated without task-specific fine-tuning. As with GPT-5.5, the base Qwen3-32B model was evaluated under both zero-shot and few-shot prompting conditions. The zero-shot prompt contained the same task definition, decision criteria, contextual information, and output requirements used for GPT-5.5, without labeled demonstrations. The few-shot condition additionally included the same four labeled training examples used in the GPT-5.5 few-shot condition.

For inference, Qwen3-32B was loaded using 4-bit NF4 quantization with double quantization enabled and bfloat16 computation. Greedy decoding was used with sampling disabled (\texttt{do\_sample=False}). Temperature was set to 0 and \texttt{top\_p} to 1.0. The maximum number of newly generated tokens was 256, the maximum input sequence length was 4,096 tokens, and the inference batch size was 4.

Both zero-shot and few-shot prompting were evaluated separately using the $\pm 1$ and $\pm 2$ boundary-instance formats. Holding the task definition, demonstrations, and contextual information constant across GPT-5.5 and the base Qwen3-32B model enabled a more direct comparison of the two model families under equivalent prompting conditions. The base-model results also provided a reference against which the contribution of task-specific Qwen3-32B fine-tuning could be assessed.

\subsubsection{Qwen3-32B Fine-Tuning}

Qwen3-32B (\texttt{Qwen/Qwen3-32B}) was adapted to the revision boundary classification task using parameter-efficient fine-tuning with LoRA. The base model was loaded using 4-bit NF4 quantization with double quantization and bfloat16 computation, following a QLoRA-style configuration. LoRA adapters were applied to the query, key, value, and output projection modules (\texttt{q\_proj}, \texttt{k\_proj}, \texttt{v\_proj}, and \texttt{o\_proj}) and the feed-forward projection modules (\texttt{gate\_proj}, \texttt{up\_proj}, and \texttt{down\_proj}). This configuration resulted in 268,435,456 trainable parameters out of 33,030,558,720 total parameters, corresponding to approximately 0.81\% of the model parameters.

Hyperparameters were tuned using the validation set, with positive-class F1 as the optimization objective. The search varied LoRA rank, LoRA alpha, LoRA dropout, learning rate, and training duration. The selected configuration used a LoRA rank of 32, an alpha of 32, a dropout rate of .05, and a learning rate of $6.4 \times 10^{-5}$. The final models were trained for up to three epochs using AdamW, with validation conducted every 100 training steps and early stopping enabled with a patience of one evaluation interval.

During fine-tuning, the supervised target was the adjudicated \texttt{same\_unit} label indicating whether the two focal operations belonged to the same MRU. At inference, the fine-tuned models were required to return the same structured output format as the prompted models, including \texttt{candidate\_boundary\_id}, \texttt{same\_unit}, \texttt{confidence}, and a brief \texttt{linguistic\_analysis}. The confidence and linguistic analysis fields were generated only at inference and were not used as supervision targets. Only the \texttt{same\_unit} prediction was used to compute the classification metrics reported in this study.

Separate LoRA adapters were trained for the two boundary instance formats: one using the $\pm 1$ context window and the other using the $\pm 2$ context window. Both were trained using the instruction-only task format without few-shot demonstrations. During inference, each adapter was loaded onto the quantized Qwen3-32B base model and evaluated using the same context window format on which it had been trained. The same greedy decoding configuration used for the base Qwen3-32B model was applied, with sampling disabled, a maximum of 256 newly generated tokens, a maximum input sequence length of 4,096 tokens, and an inference batch size of 4. Fine-tuning was conducted on a supercomputer using an NVIDIA A100 GPU with 80 GB of memory.

This design enabled task-specific adaptation to be evaluated by comparing each fine-tuned Qwen3-32B model with the corresponding zero-shot base Qwen3-32B condition while holding the underlying model architecture and boundary-instance representation constant.

\subsection{Post-Processing Procedure}
\label{sec:postprocessing}

Following model inference, a deterministic post-processing procedure was applied to enforce a structural constraint involving \texttt{preserve} operations that was also considered during human annotation. Unlike the proximity heuristic, which generated predictions directly from the operation types of the focal pair, post-processing operated on existing LLM predictions and conditionally modified them based on the surrounding boundary structure. The procedure was applied independently within each draft pair while preserving the original ordering of candidate boundaries.

Post-processing was implemented as a single-pass correction based exclusively on the original model predictions. For each candidate boundary originally predicted as \texttt{False}, the immediately following boundary was changed from \texttt{True} to \texttt{False} when the left operation of that following boundary was a \texttt{preserve} operation. Similarly, the immediately preceding boundary was changed from \texttt{True} to \texttt{False} when the right operation of that preceding boundary was a \texttt{preserve} operation. Predictions already labeled \texttt{False} remained unchanged, and the procedure never converted a \texttt{False} prediction to \texttt{True}.

Importantly, boundaries changed during post-processing were not treated as new triggers for additional corrections. Thus, the rule did not propagate recursively through the operation sequence. No post-processing was applied across draft-pair boundaries. This single-pass procedure was intended to apply the same local boundary checking constraint used during human annotation without allowing corrections to propagate beyond the immediate context of an original model-predicted boundary.

For each LLM condition, both the original model predictions and the post-processed predictions were retained and evaluated separately so that the contribution of the deterministic rule could be distinguished from that of the LLM itself. Post-processing was not applied to the two non-LLM baselines.

\subsection{Evaluation Measures}

Predictions were compared with the adjudicated gold-standard annotations using accuracy, precision, recall, F1 score, and macro-F1. Metrics were computed using scikit-learn. For class-specific measures, \texttt{same\_unit = True} was treated as the positive class.

Accuracy represented the proportion of all candidate boundaries classified correctly. Precision represented the proportion of predicted same unit boundaries that were labeled as same unit instances in the gold-standard data. Recall represented the proportion of gold-standard same unit instances correctly identified. Positive-class F1 was calculated as the harmonic mean of precision and recall. Macro-F1 was calculated as the unweighted mean of the F1 scores for the positive and negative classes.

Because the boundary data were imbalanced toward the negative class, macro-F1 was used as the primary aggregate performance measure, while accuracy was reported as a complementary measure. Precision, recall, and F1 for the positive class were additionally reported to characterize the ability of each approach to identify adjacent operations that should be combined into the same MRU. The majority and proximity baselines were evaluated once on the held-out test set. LLM performance measures were calculated separately for raw and post-processed predictions.

\subsection{AI Use Disclosure}

Generative AI systems were used as research objects and analytic tools as described in the Methods section. The authors take full responsibility for the accuracy, integrity, and final content of the manuscript.

\section{Results}

Performance was evaluated on the held-out test set. Two non-LLM baselines were evaluated alongside GPT-5.5, the base Qwen3-32B model, and the fine-tuned Qwen3-32B models. GPT-5.5 and the base Qwen3-32B model were each evaluated under zero-shot and few-shot prompting with $\pm 1$ and $\pm 2$ context windows, whereas the fine-tuned Qwen3-32B models were evaluated using instruction-only prompts. LLM performance was assessed both before and after deterministic post-processing. Because the test set was imbalanced toward the negative class, macro-F1 was treated as the primary aggregate performance measure.

\subsection{Baseline and Raw Model Performance}

Table~\ref{tab:before-postprocessing} presents the two non-LLM baselines together with the raw LLM predictions. The majority baseline achieved an accuracy of .825 because of the substantial class imbalance, but it failed to identify any positive same unit boundaries and produced a macro-F1 of only .452. In contrast, the proximity heuristic performed strongly, achieving .912 accuracy, .852 precision, .597 recall, .702 positive-class F1, and .825 macro-F1.

\begin{table}[htbp]
\centering
\caption{Baseline and LLM Performance Before Post-Processing}
\label{tab:before-postprocessing}
\resizebox{\textwidth}{!}{%
\begin{tabular}{lllrrrrr}
\toprule
\textbf{Model} & \textbf{Prompting condition} & \textbf{Context window} & \textbf{Accuracy} & \textbf{Precision} & \textbf{Recall} & \textbf{F1} & \textbf{Macro-F1} \\
\midrule
Majority baseline & -- & -- & .825 & .000 & .000 & .000 & .452 \\
Proximity heuristic & -- & -- & .912 & .852 & .597 & .702 & .825 \\
\midrule
GPT-5.5 & Zero-shot & $\pm 1$ & .522 & .246 & .844 & .381 & .496 \\
GPT-5.5 & Few-shot & $\pm 1$ & .626 & .292 & .805 & .429 & .575 \\
GPT-5.5 & Zero-shot & $\pm 2$ & .574 & .273 & \textbf{.870} & .416 & .540 \\
GPT-5.5 & Few-shot & $\pm 2$ & .628 & .288 & .766 & .418 & .573 \\
Qwen3-32B Base Model & Zero-shot & $\pm 1$ & .637 & .296 & .779 & .429 & .581 \\
Qwen3-32B Base Model & Few-shot & $\pm 1$ & .650 & .299 & .763 & .430 & .589 \\
Qwen3-32B Base Model & Zero-shot & $\pm 2$ & .620 & .290 & .805 & .426 & .571 \\
Qwen3-32B Base Model & Few-shot & $\pm 2$ & .713 & .344 & .701 & .462 & .633 \\
Qwen3-32B fine-tuned & Instruction-only & $\pm 1$ & .893 & .778 & .545 & .641 & .789 \\
Qwen3-32B fine-tuned & Instruction-only & $\pm 2$ & \textbf{.925} & \textbf{.855} & .688 & \textbf{.763} & \textbf{.859} \\
\bottomrule
\end{tabular}%
}
\end{table}

The proximity heuristic outperformed every prompted LLM condition by a substantial margin in macro-F1 and also exceeded the fine-tuned Qwen3-32B model using the $\pm 1$ representation. Only the fine-tuned Qwen3-32B model trained and evaluated with the $\pm 2$ representation exceeded the heuristic, achieving a macro-F1 of .859 compared with .825. Notably, the two approaches achieved nearly identical precision (.855 for the fine-tuned model and .852 for the heuristic), whereas the fine-tuned model achieved higher recall (.688 vs. .597). Thus, the performance advantage of the strongest fine-tuned configuration was primarily associated with identifying additional positive same unit boundaries rather than reducing false positive predictions.

Among the prompted LLMs, few-shot prompting generally improved accuracy and macro-F1 relative to zero-shot prompting, although recall tended to decrease. The strongest raw prompted condition was the base Qwen3-32B model under few-shot prompting with the $\pm 2$ context window (macro-F1 = .633), which nevertheless remained well below the proximity heuristic. The raw GPT-5.5 and base Qwen3-32B models generally favored recall over precision, indicating a tendency to predict same unit relationships more frequently than the heuristic and fine-tuned models.

Context-window effects varied across LLM conditions. Increasing the window from $\pm 1$ to $\pm 2$ improved zero-shot GPT-5.5 performance but had little effect under few-shot prompting. For the base Qwen3-32B model, the larger context window was most beneficial under few-shot prompting. The clearest difference was observed between the two fine-tuned configurations, with the model trained and evaluated using the $\pm 2$ representation achieving higher accuracy, recall, F1, and macro-F1 than the corresponding $\pm 1$ model.

\subsection{Performance After Post-Processing}

Table~\ref{tab:after-postprocessing} presents LLM performance after application of the deterministic post-processing procedure. Post-processing generally increased precision and reduced recall, consistent with the fact that the procedure could change positive predictions to negative predictions but not the reverse.

\begin{table}[htbp]
\centering
\caption{LLM Performance After Post-Processing}
\label{tab:after-postprocessing}
\resizebox{\textwidth}{!}{%
\begin{tabular}{lllrrrrr}
\toprule
\textbf{Model} & \textbf{Prompting condition} & \textbf{Context window} & \textbf{Accuracy} & \textbf{Precision} & \textbf{Recall} & \textbf{F1} & \textbf{Macro-F1} \\
\midrule
GPT-5.5 & Zero-shot & $\pm 1$ & .853 & .558 & \textbf{.753} & .641 & .774 \\
GPT-5.5 & Few-shot & $\pm 1$ & .884 & .651 & .727 & .687 & .808 \\
GPT-5.5 & Zero-shot & $\pm 2$ & .868 & .598 & \textbf{.753} & .667 & .792 \\
GPT-5.5 & Few-shot & $\pm 2$ & .868 & .623 & .623 & .623 & .772 \\
Qwen3-32B Base Model & Zero-shot & $\pm 1$ & .839 & .529 & .701 & .603 & .751 \\
Qwen3-32B Base Model & Few-shot & $\pm 1$ & .845 & .543 & .671 & .600 & .752 \\
Qwen3-32B Base Model & Zero-shot & $\pm 2$ & .857 & .578 & .675 & .623 & .767 \\
Qwen3-32B Base Model & Few-shot & $\pm 2$ & .875 & .662 & .584 & .621 & .773 \\
Qwen3-32B fine-tuned & Instruction-only & $\pm 1$ & .902 & .870 & .519 & .650 & .797 \\
Qwen3-32B fine-tuned & Instruction-only & $\pm 2$ & \textbf{.921} & \textbf{.875} & .636 & \textbf{.737} & \textbf{.845} \\
\bottomrule
\end{tabular}%
}
\end{table}

Post-processing produced marked improvements for GPT-5.5 and the base Qwen3-32B model, largely through increases in precision. However, even after post-processing, none of the prompted LLM conditions exceeded the proximity heuristic's macro-F1 of .825; the strongest prompted condition was GPT-5.5 under few-shot prompting with the $\pm 1$ window (macro-F1 = .808). The post-processed fine-tuned $\pm 1$ model also remained below the heuristic (macro-F1 = .797).

The fine-tuned Qwen3-32B model using the $\pm 2$ representation remained the strongest overall LLM condition after post-processing, with a macro-F1 of .845. However, this was lower than its raw macro-F1 of .859. Thus, only the fine-tuned $\pm 2$ configuration exceeded the proximity heuristic in macro-F1, both before and after post-processing, and its strongest performance was obtained without post-processing.

\section{Discussion}

This study evaluated whether LLMs can make accurate boundary judgments in structured writing revision data and whether they provide value beyond simpler non-LLM baselines. The findings revealed that general purpose prompting alone was not sufficient: both GPT-5.5 and the base Qwen3-32B model were outperformed by a simple proximity heuristic that used only TRACE operation types. However, task-specific fine-tuning enabled Qwen3-32B to exceed the heuristic under the strongest configuration. These results suggest that LLMs can support MRU identification, but their effectiveness depends strongly on task adaptation and should be evaluated against transparent structural baselines rather than assumed from model capability alone.

With respect to RQ1, the two non-LLM baselines provided important reference points for interpreting LLM performance. The majority baseline achieved relatively high accuracy because of class imbalance, but it failed to identify any positive same unit boundaries and produced a low macro-F1. In contrast, the proximity heuristic performed strongly, achieving a macro-F1 of .825 by predicting that adjacent non-preserve operations belonged to the same MRU and that boundaries involving a \texttt{preserve} operation separated units. This result indicates that MRU boundaries in the present data exhibit substantial structural regularity and that operation type alone provides considerable information about whether adjacent operations should be grouped.

The comparison between the proximity heuristic and the LLM-based approaches is central to understanding the capability of LLMs in this task. Although the prompted LLMs received source and revised sentence context, neighboring revision operations, task instructions, and in some conditions labeled examples, they did not outperform a deterministic rule based only on the operation types of the focal pair. The raw prompted models generally achieved high recall but low precision, reflecting a tendency to merge adjacent operations too frequently. Thus, general purpose LLMs appeared sensitive to possible relationships between neighboring edits, but they were less effective at applying the structural constraints needed to avoid over-merging. This finding cautions against assuming that model scale or linguistic fluency will automatically produce reliable judgments in structured revision analytics.

At the same time, the strongest fine-tuned model did provide additional value beyond the heuristic. The fine-tuned Qwen3-32B model trained and evaluated with the $\pm 2$ representation achieved a macro-F1 of .859, compared with .825 for the proximity heuristic. The magnitude of this gain was modest, but the precision--recall profiles clarify where the improvement occurred. Precision was nearly identical for the fine-tuned model (.855) and the heuristic (.852), whereas recall increased from .597 to .688. This pattern suggests that fine-tuning helped the model identify additional same unit relationships that could not be captured from operation type alone while largely preserving the heuristic's ability to avoid false positive merges. Such cases may include MRUs in which a \texttt{preserve} operation occurs internally between related revisions or cases in which operation type alone is insufficient to determine the functional relationship between neighboring changes.

With respect to RQ2, performance among the LLM-based approaches varied meaningfully by model configuration, prompting strategy, and local context window size. Task-specific fine-tuning was the clearest source of improvement, outperforming both GPT-5.5 prompting and base Qwen3-32B prompting. This finding is consistent with recent revision classification research showing that LLM adaptation can improve performance on nuanced edit-level classification tasks \parencite{ruan2024large, liu2025efficient}. The present study extends that line of work to an earlier stage of the revision analysis pipeline: before revision intent can be classified, granular operations must first be grouped into meaningful units.

The effects of prompting and context size were more limited and less consistent. Few-shot prompting generally improved accuracy and macro-F1 relative to zero-shot prompting, but the gains were modest and often accompanied by reductions in recall. More importantly, neither zero-shot nor few-shot prompting approached the performance of the proximity heuristic. This suggests that a small number of demonstrations was insufficient for the prompted models to consistently exploit the structural regularities of the task. The larger $\pm 2$ context window was most beneficial for the fine-tuned model, but its effects were mixed for the prompted models. Thus, providing more contextual information was not sufficient by itself. The model also needed to be adapted to use that information in a manner aligned with the target boundary judgment.

With respect to RQ3, deterministic post-processing improved the non-fine-tuned LLMs but provided little additional benefit after fine-tuning. Post-processing substantially increased precision for GPT-5.5 and the base Qwen3-32B model by reducing false positive same unit predictions. This pattern suggests that some weaknesses of prompted LLMs could be corrected by explicitly enforcing known task constraints rather than relying on the models to infer those constraints from instructions and examples alone. In contrast, post-processing slightly reduced macro-F1 for the strongest fine-tuned configuration, suggesting that fine-tuning had already helped the model learn some of the structural regularities that the deterministic rule was designed to enforce.

Taken together, these findings suggest a complementary relationship between transparent rules and task-adapted models. Simple structural heuristics may provide strong performance when reliable regularities are available in the revision representation. However, task-specific model adaptation may recover additional meaningful relationships when operation type alone is insufficient. This balance is important for revision analytics because the way granular operations are grouped determines what counts as a revision unit in subsequent analyses. More accurate MRU identification can therefore support more interpretable measures of revision frequency, scope, and purpose.

More broadly, MRU identification is not only a modeling problem but also a representational problem for learning analytics. The study shows that moving from granular operations to meaningful units requires attention to both the structure of the computational trace and the interpretive judgment needed to connect operations into purposeful learner actions. Although the present evaluation used TRACE-generated operations, similar representational challenges arise in other forms of writing process and revision data, including keystroke logs, where low-level actions must be aggregated into interpretable events \parencite{tian2026keystroke}. The results therefore support a selective approach to LLM-based revision analytics: general purpose prompting may be insufficient, but task-adapted models can add value when evaluated against strong and transparent baselines.

\subsection{Limitations and Future Work}

Several limitations should be considered when interpreting these findings. First, the dataset consisted of narrative essays produced by undergraduate students within a single institutional context, which may limit the generalizability of the results to other genres, educational levels, and writing contexts. The strong performance of the proximity heuristic may also depend partly on the structural properties of TRACE operations. Future research should therefore examine whether similar structural regularities arise with other revision representations and datasets.

Second, the non-LLM comparisons were intentionally limited to simple baselines designed to test class imbalance and operation-level structural regularities. The study did not evaluate a learned non-LLM classifier such as BERT, RoBERTa, or a feature-based supervised model. The present findings therefore establish that fine-tuned Qwen3-32B can outperform the majority and proximity baselines under the strongest configuration, but they do not establish an advantage over conventional supervised language models. Future work should compare task-adapted LLMs with smaller learned models to determine whether the additional scale and generative capacity of an LLM provide benefits beyond supervised contextual classification.

Third, the few-shot experiments used a single fixed set of four demonstrations. The observed prompting effects may therefore be sensitive to demonstration selection. Future studies could evaluate multiple demonstration sets, alternative selection strategies, or larger numbers of examples to determine whether the patterns observed here are robust.

Finally, although the goal of the task is to recover MRUs, evaluation was conducted at the individual boundary level. Boundary-level performance does not fully capture the quality of the resulting multi-operation units because a single boundary error can split a coherent MRU or merge otherwise distinct MRUs. Future work should therefore complement boundary-classification metrics with unit- or sequence-level evaluation and examine how segmentation errors propagate to downstream analyses of revision behavior.

\section{Conclusion}

This study positioned MRU identification as a boundary-judgment problem in revision analytics and evaluated whether LLM-based approaches can support that judgment more effectively than simple structural baselines. The results provide a nuanced assessment of LLM capability in structured revision analytics. Prompting general purpose LLMs, even with task instructions, examples, and local revision context, did not outperform a proximity-based heuristic that relied only on TRACE operation types. This finding underscores the importance of comparing LLM-based methods with transparent baselines that reflect the structure of the data.

Task-specific fine-tuning nevertheless provided additional value. The fine-tuned Qwen3-32B model trained and evaluated with the $\pm 2$ contextual representation achieved the strongest performance, with a macro-F1 of .859 compared with .825 for the proximity heuristic. The improvement was driven primarily by higher recall while maintaining comparable precision, suggesting that task adaptation can help identify meaningful revision unit relationships that are not captured by operation type alone. Deterministic post-processing also substantially improved the non-fine-tuned LLMs, further demonstrating the value of explicit task knowledge when applying LLMs to structured revision data.

Together, these findings support a selective approach to LLM-based revision analytics. Simple, transparent rules may be sufficient when MRU boundaries align closely with structural regularities in the revision representation, whereas task-specific model adaptation can provide additional value when revision-unit decisions require contextual interpretation. More broadly, the study highlights the importance of evaluating computationally complex models against meaningful structural baselines when transforming granular learner traces into interpretable units of analysis.

\section*{Acknowledgements}

The research reported here was supported by the Institute of Education Sciences, U.S. Department of Education, through Grants R305A180261, R305N210041, and R305T240035 to Arizona State University. The opinions expressed are those of the authors and do not represent views of the Institute or the U.S. Department of Education.

\printbibliography

@unpublished{tian-inprep-feedback,
  author = {Tian, Yu and Potter, Andrew and Christhilf, Katerina and Qi, Sun and Early, Jessica and Graham, Steve and McNamara, Danielle},
  title = {Effects of Prompting Strategies for {AI}-Generated Writing Feedback on Student Writing Outcomes},
  note = {Manuscript in preparation}
}

@inproceedings{tian2026trace,
  author    = {Tian, Yu and Christhilf, Katerina and Potter, Andrew and Early, Jessica and Graham, Steve and McNamara, Danielle},
  title     = {{TRACE}: Automated Multi-Granularity Analysis of Text Revision},
  booktitle = {Proceedings of {AIME-CON} 2026},
  year      = {2026}
}

@article{vach2023gwet,
  author = {Vach, Werner and Gerke, Oke},
  title = {Gwet's AC1 is not a substitute for Cohen's kappa: A comparison of basic properties},
  journal = {MethodsX},
  volume = {10},
  pages = {102212},
  year = {2023},
  doi = {10.1016/j.mex.2023.102212}
}

@article{fitzgerald1987research,
  title={Research on revision in writing},
  author={Fitzgerald, Jill},
  journal={Review of educational research},
  volume={57},
  number={4},
  pages={481--506},
  year={1987},
  publisher={Sage Publications Sage CA: Thousand Oaks, CA}
}

@article{flower1986detection,
  title={Detection, diagnosis, and the strategies of revision},
  author={Flower, Linda and Hayes, John R and Carey, Linda and Schriver, Karen and Stratman, James},
  journal={College Composition \& Communication},
  volume={37},
  number={1},
  pages={16--55},
  year={1986},
  publisher={NCTE}
}

@article{conijn2024automated,
  title={Automated extraction of revision events from keystroke data},
  author={Conijn, Rianne and Dux Speltz, Emily and Chukharev-Hudilainen, Evgeny},
  journal={Reading and Writing},
  volume={37},
  number={2},
  pages={483--508},
  year={2024},
  publisher={Springer}
}

@article{tian2025exploring,
  title={Exploring the application of keystroke logging techniques to research in second language (L2) writing},
  author={Tian, Yu and Cushing, Sara T},
  journal={Research Methods in Applied Linguistics},
  volume={4},
  number={1},
  pages={100179},
  year={2025},
  publisher={Elsevier}
}

@article{limpo2014children,
  title={Children's high-level writing skills: Development of planning and revising and their contribution to writing quality},
  author={Limpo, Teresa and Alves, Rui A and Fidalgo, Raquel},
  journal={British Journal of Educational Psychology},
  volume={84},
  number={2},
  pages={177--193},
  year={2014},
  publisher={Wiley Online Library}
}

@article{zhu2020effect,
  title={The effect of automated feedback on revision behavior and learning gains in formative assessment of scientific argument writing},
  author={Zhu, Mengxiao and Liu, Ou Lydia and Lee, Hee-Sun},
  journal={Computers \& Education},
  volume={143},
  pages={103668},
  year={2020},
  publisher={Elsevier}
}

@inproceedings{jiang2022arxivedits,
  title={arxivedits: Understanding the human revision process in scientific writing},
  author={Jiang, Chao and Xu, Wei and Stevens, Samuel},
  booktitle={Proceedings of the 2022 Conference on Empirical Methods in Natural Language Processing},
  pages={9420--9435},
  year={2022}
}

@inproceedings{swanson2012correction,
  title={Correction detection and error type selection as an ESL educational aid},
  author={Swanson, Ben and Yamangil, Elif},
  booktitle={Proceedings of the 2012 Conference of the North American Chapter of the Association for Computational Linguistics: Human Language Technologies},
  pages={357--361},
  year={2012}
}

@inproceedings{xue2014improved,
  title={Improved correction detection in revised ESL sentences},
  author={Xue, Huichao and Hwa, Rebecca},
  booktitle={Proceedings of the 52nd Annual Meeting of the Association for Computational Linguistics (Volume 2: Short Papers)},
  pages={599--604},
  year={2014}
}

@inproceedings{felice2016automatic,
  title={Automatic extraction of learner errors in ESL sentences using linguistically enhanced alignments},
  author={Felice, Mariano and Bryant, Christopher and Briscoe, Ted},
  booktitle={Proceedings of COLING 2016, the 26th International Conference on Computational Linguistics: Technical Papers},
  pages={825--835},
  year={2016}
}

@inproceedings{zhang2015annotation,
  title={Annotation and classification of argumentative writing revisions},
  author={Zhang, Fan and Litman, Diane},
  booktitle={Proceedings of the tenth workshop on innovative use of NLP for building educational applications},
  pages={133--143},
  year={2015}
}

@inproceedings{lan2026making,
  title={Making Revisions Understandable: A Survey of Edit Intentions, Methods, and Applications},
  author={Lan, Fangping and Zhang, Qi and Dragut, Eduard},
  booktitle={Findings of the Association for Computational Linguistics: ACL 2026},
  pages={35003--35019},
  year={2026}
}

@inproceedings{nebel2025towards,
  title={Towards automated characterization of revision events in student writing},
  author={Nebel, L{\'e}o and Bouchet, Fran{\c{c}}ois and Luengo, Vanda and Couraud, Mathilde},
  booktitle={European Conference on Technology Enhanced Learning},
  pages={397--411},
  year={2025},
  organization={Springer}
}

@inproceedings{bryant2017automatic,
  title={Automatic annotation and evaluation of error types for grammatical error correction},
  author={Bryant, Christopher and Felice, Mariano and Briscoe, Ted},
  booktitle={Proceedings of the 55th annual meeting of the association for computational linguistics (Volume 1: Long Papers)},
  pages={793--805},
  year={2017}
}

@inproceedings{ruan2024large,
  title={Are large language models good classifiers? a study on edit intent classification in scientific document revisions},
  author={Ruan, Qian and Kuznetsov, Ilia and Gurevych, Iryna},
  booktitle={Proceedings of the 2024 conference on empirical methods in natural language processing},
  pages={15049--15067},
  year={2024}
}

@inproceedings{liu2025efficient,
  title={Efficient Layer-wise LLM Fine-tuning for Revision Intention Prediction},
  author={Liu, Zhexiong and Litman, Diane},
  year={2025},
  organization={Association for Computational Linguistics}
}

@incollection{tian2026keystroke,
  title={Keystroke logging},
  author={Tian, Yu and Crossley, Scott},
  booktitle={Digital and Internet-Based Research Methods in Applied Linguistics},
  pages={310--335},
  year={2026},
  publisher={John Benjamins Publishing Company}
}

@article{mouchel2023understanding,
  title={Understanding revision behavior in adaptive writing support systems for education},
  author={Mouchel, Luca and Wambsganss, Thiemo and Mejia-Domenzain, Paola and K{\"a}ser, Tanja},
  journal={arXiv preprint arXiv:2306.10304},
  year={2023}
}

\end{document}